%% file: main.tex
\documentclass[10pt,twocolumn,letterpaper]{article}

\usepackage{iccv}              


\usepackage{multirow}
\usepackage{graphicx} 
\usepackage{lscape}   
\usepackage{array}    
\usepackage{algorithm}
\usepackage{algpseudocode}

\newcommand{\ourdataset}{\textsc{InterDrive}\xspace}

\newcommand{\ourmethod}{\textsc{LangTraj}\xspace}

\definecolor{iccvblue}{rgb}{0.21,0.49,0.74}
\usepackage[pagebackref,breaklinks,colorlinks,allcolors=iccvblue]{hyperref}

\def\paperID{7032} 
\def\confName{ICCV}
\def\confYear{2025}

\title{\ourmethod: Diffusion Model and Dataset for Language-Conditioned \\ Trajectory Simulation}

\author{ Wei-Jer Chang$^1$ \and
Wei Zhan$^1$ \and
Masayoshi Tomizuka$^1$ \and
Manmohan Chandraker$^{2,3}$ \and
Francesco Pittaluga$^2$ \\\\
$^1$ UC Berkeley \hspace{0.5cm} $^2$ NEC Labs America \hspace{0.5cm} $^3$ UC San Diego
}

\begin{document}

\twocolumn[{%
\renewcommand\twocolumn[1][]{#1}%
\maketitle
\begin{center}
\fbox{\includegraphics[width=\textwidth]{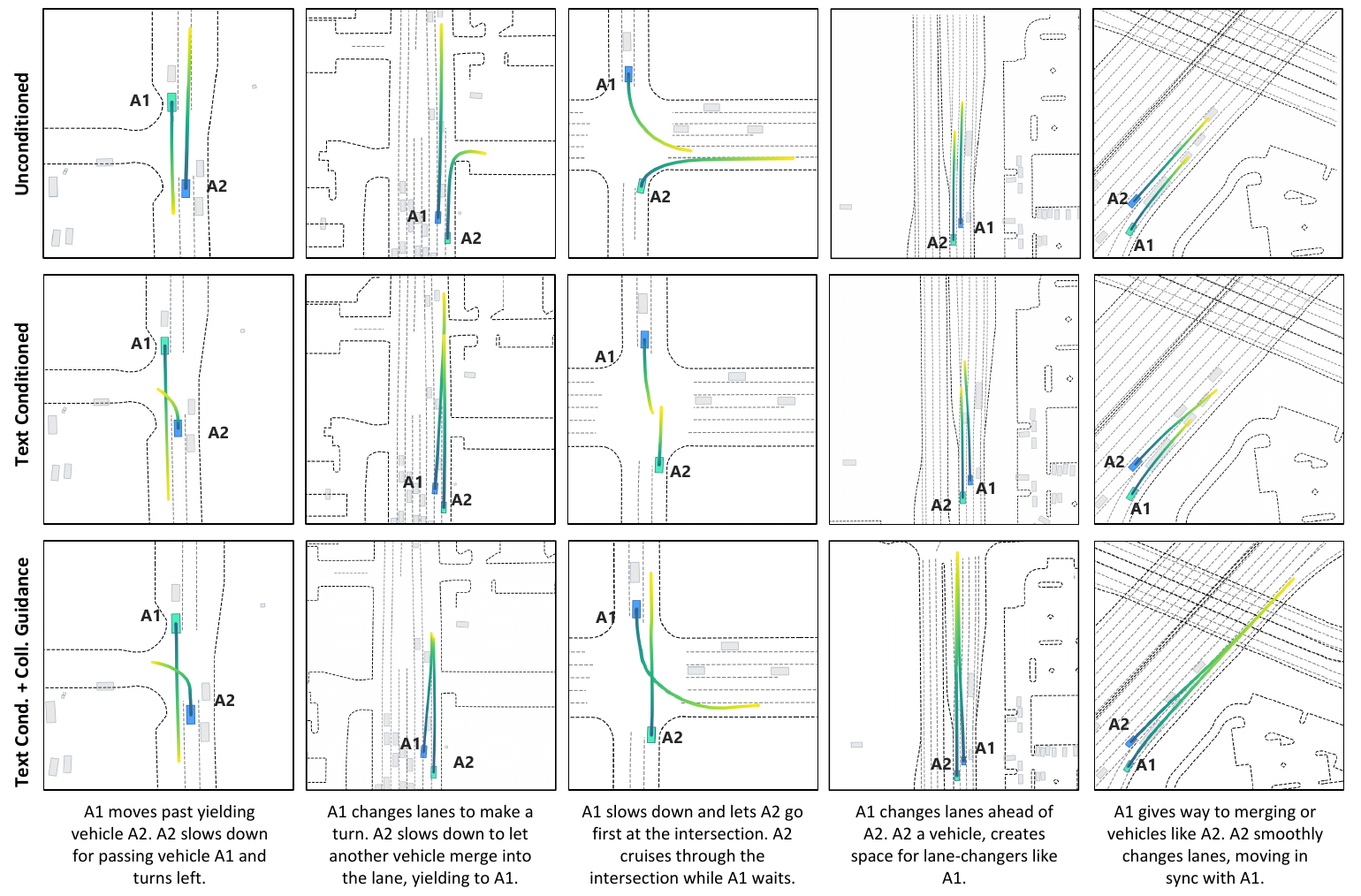}}
\captionof{figure}{\textbf{Unconditioned (top), Text-Conditioned (mid), and Safety-Critical + Text-Conditioned (bot) Simulations by \ourmethod.} Adversarial collision guidance is applied in conjunction with text conditioning to generate the safety-critical scenarios shown in the bottom row, where the dark blue car serves as the adversarial agent. Language annotations are from the test set of \ourdataset.}
\label{teaser}
\end{center}
}]

\input{sec/0_abstract}    

\input{sec/1_intro}
\input{sec/2_rw}
\input{sec/3_method}

\input{sec/4_exp}

\input{sec/5_conc}

\input{sec/ack}

{
    \small
    \bibliographystyle{ieeenat_fullname}
    \bibliography{main}
}

\input{sec/X_suppl}



\end{document}

%% file: sec/0_abstract.tex

\begin{abstract}
Evaluating autonomous vehicles with controllability allows for scalable testing in counterfactual or structured settings, improving both efficiency and safety.  We introduce \ourmethod, a language-conditioned scene-diffusion model that simulates the joint behavior of all agents in traffic scenarios. By conditioning on natural language inputs, \ourmethod enables flexible and intuitive control over interactive behaviors, generating nuanced and realistic scenarios. Unlike prior approaches that rely on domain-specific guidance functions, \ourmethod incorporates language conditioning during training for more intuitive traffic simulation control. In addition, we propose a novel closed-loop training strategy for diffusion models to enhance realism in closed-loop simulation. To support language-conditioned simulation, we develop a scalable pipeline for annotating agent-agent interactions and single-agent behaviors, which we use to develop \ourdataset, a large-scale dataset offering diverse and interactive labels for training language-conditioned diffusion models. Validated on the Waymo Open Motion Dataset, \ourmethod demonstrates strong performance in both realism, language controllability, and language-conditioned safety-critical simulation, establishing a new paradigm for flexible and scalable autonomous vehicle testing. Project website: https://langtraj.github.io/.

\end{abstract}

%% file: sec/1_intro.tex
\section{Introduction}
\label{sec:intro}  

Traffic simulation is essential for the safe and scalable development of autonomous vehicles (AVs). By simulating interactions among multiple traffic participants, it enables AVs to handle a wide range of realistic scenarios. This approach is critical for three main reasons: 1) it accelerates development by enabling large-scale, repeatable testing in controlled environments, 2) it provides structured testing to validate vehicle behavior across diverse conditions, and 3) it allows AVs to train for real-world complexities, improving safety and reliability. A key aspect of simulation is \textit{controllability}: the ability to model the interactive behavior of other road users, which AVs must learn to navigate and respond to safely and effectively.

Traditionally, structured testing of AVs has relied on manually designed scenarios to simulate failure cases or counterfactual situations, such as collisions or typical interactive behaviors. While effective for targeted testing, this approach is inherently limited in scalability. Recent advances in diffusion-based generative models demonstrate strong capabilities in simulating complex distributions with flexible controllability \cite{CTG,chang2025safesim}, though this often depends on domain-specific heuristic guidance functions constructed based on human knowledge \textit{post-training}. Our key insight is to \textbf{leverage language conditioning during training} to directly learn semantics from the data distribution, enabling users—including non-experts—to generate counterfactual scenarios with ease. By conditioning on natural language, we can significantly expand the range of possible scenarios, creating a more capable model as the data scales and allowing for diverse driving behaviors. Note that direct conditioning on language is orthogonal to guidance functions, allowing us to combine the strengths of both human knowledge and data-driven insights for more diverse simulations.


We present \ourmethod, a language-controlled diffusion model that generates realistic and controllable trajectories for AV simulation by conditioning on natural language prompts. This approach enables \ourmethod to respond to diverse instructions such as “yield to the right” or “merge left,” capturing complex interactive behaviors for counterfactual testing across varied driving scenarios.

Achieving this functionality requires overcoming two key challenges:  developing a model that can effectively condition on diverse user inputs during closed-loop simulation and acquiring high-quality data with natural language labels. To address these challenges, we first introduce a scene-diffusion model, which jointly models multi-agent behaviors while conditioning on text inputs, enabling flexible and scalable AV testing. We propose a novel \textit{closed-loop training strategy} for diffusion models to improve closed-loop realism and reduce error accumulation during iterative rollouts. Contrary to prior diffusion model works \cite{CTG, jiang2024scenediffuser,chang2025safesim} that adopt inference techniques such as constraint and guidance methods to steer predictions, which can slow inference and require careful balancing of multiple objectives, we introduce a closed-loop training strategy that explicitly enhances model stability and realism during closed-loop simulation. Additionally, guidance methods can still be applied post-training to refine behavior further if needed.

To enable language-conditioned simulation, we introduce \ourdataset, a comprehensive dataset with 150k human-labeled annotations, specifically focusing on \textit{interactive} agent-agent behaviors (e.g., merging, yielding, and passing) with rich language annotations. In addition, we include heuristic-based labels for single-agent behaviors such as directional intent and lane changes. This dataset ensures high-quality annotations for interaction modeling, providing the necessary diversity and richness for training models that capture both interactive and individualistic driving behaviors.

Our contributions are as follows:
\begin{itemize}
    \item \textbf{\ourmethod: The First Diffusion Model Directly Conditioned on Language for Interactive Simulation.}  
    We introduce \ourmethod, the first diffusion-based trajectory generation model that directly conditions on natural language inputs for interactive simulation.

    \item \textbf{Novel Closed-Loop Training Strategy for Diffusion Models.}  
    We propose a novel closed-loop training strategy explicitly tailored for diffusion models to enhance stability and realism during closed-loop simulation.

    \item \textbf{\ourdataset~Dataset.}  
    We present \ourdataset, a new dataset of 150k human-labeled interactive traffic scenarios, supplemented with heuristically labeled single-agent behaviors, enabling scalable training of language-conditioned simulation models.
\end{itemize}




%% file: sec/2_rw.tex
\begin{figure*}
    \centering
    \includegraphics[width=\textwidth]{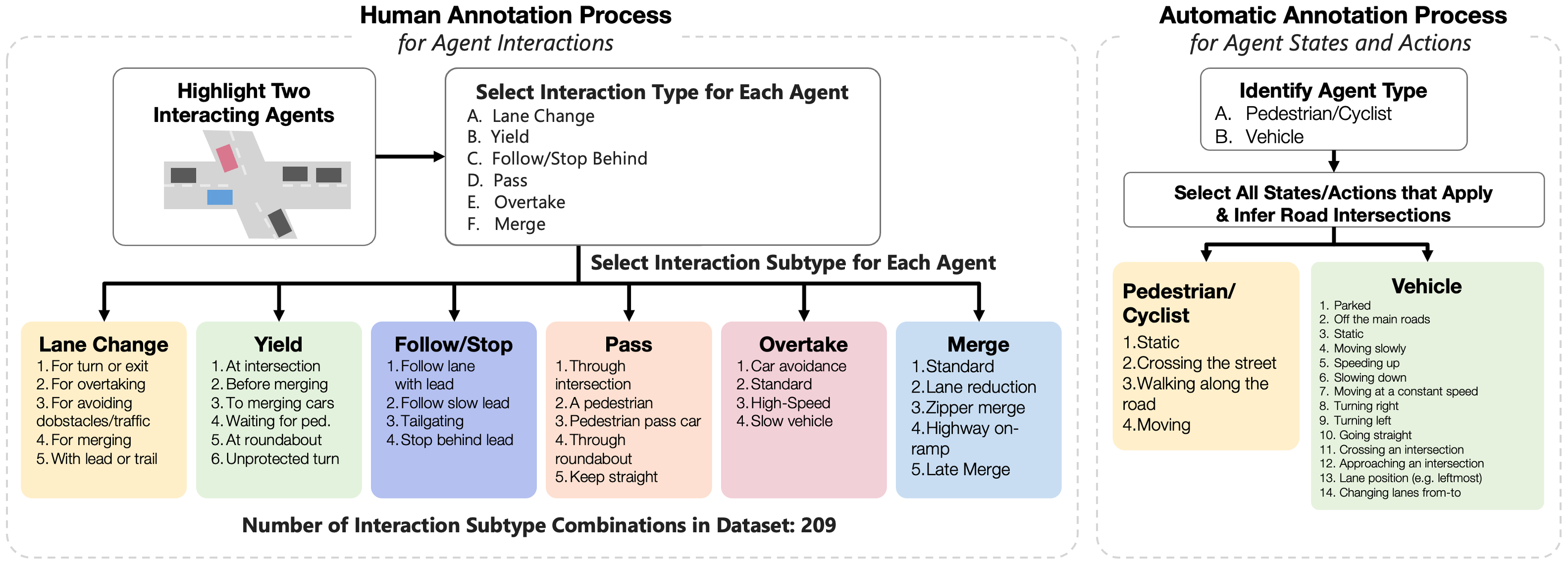} 
    \includegraphics[width=\textwidth]{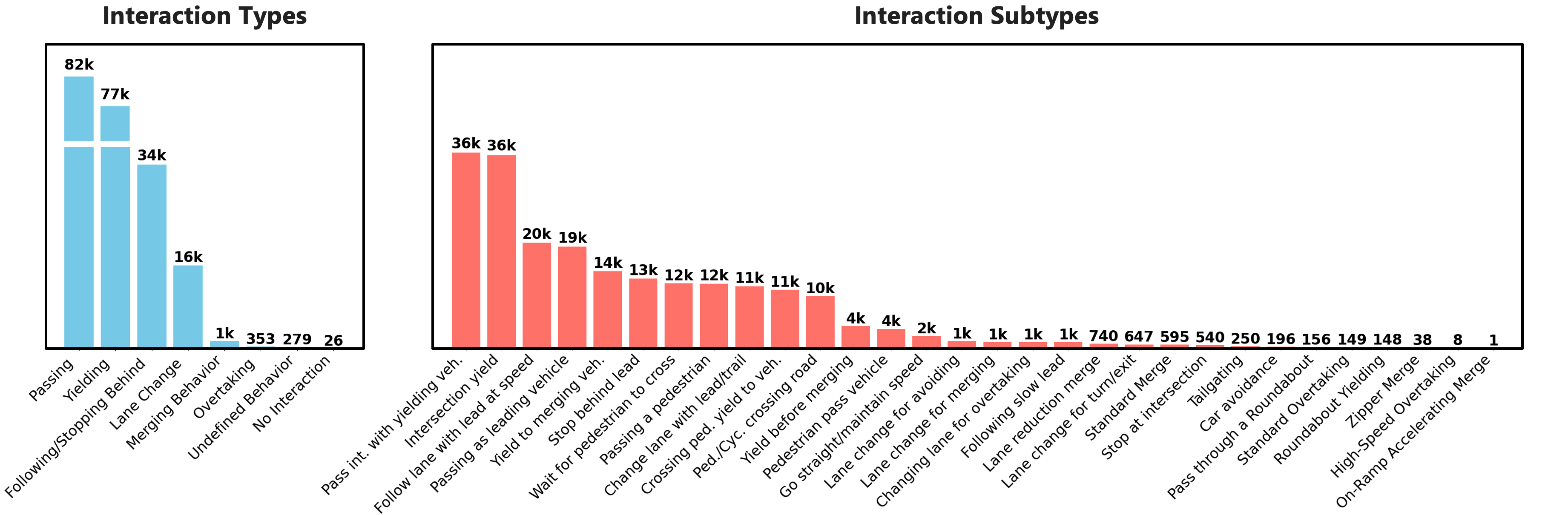} 
    \caption{\textbf{Overview of \ourdataset dataset.} \ourdataset captures nuanced agent-agent interactions in real-world driving contexts. It includes human-labeled traffic interaction annotations from Waymo Motion and NuPlan datasets, along with single-agent behavioral labels generated through heuristic annotations, offering a comprehensive view of agent actions and interactions in diverse traffic scenarios. The top part of the figure shows the annotation processes for the human and heuristic annotations. The bottom part of the figure shows the counts of each interaction and interaction subtypes in \ourdataset.}
    \label{fig:dataset} 
\end{figure*}


\section{Related Work}
\label{sec:rw}

\subsection{Traffic Simulation and Controllable Diffusion}  
Traffic simulation methods are either heuristic-based or learning-based. Heuristic models like IDM~\cite{treiber2000congested} rely on predefined rules but struggle with real-world accuracy. Learning-based approaches~\cite{suo2021trafficsim,BITS,suo2023mixsim, philion2024trajeglish} leverage real-world data for more natural behavior. TrafficSim~\cite{suo2021trafficsim} employs a variational autoencoder for scene-level simulation, while BITS~\cite{BITS} combines goal inference with imitation learning. The Waymo SimAgents challenge~\cite{waymo_sim_agent} highlights the need for realistic driving distributions, but controllable scenario generation remains underexplored.

Controllable diffusion models have advanced traffic simulation~\cite{jiang2024scenediffuser,motiondiffuser,CTG,ctg++,chang2025safesim}. Diffusion-ES~\cite{diffusion_es} combines evolutionary search with diffusion models. CTG~\cite{CTG} introduces test-time guidance, while SAFE-SIM~\cite{chang2025safesim} models adversarial behaviors. CTG++~\cite{ctg++} enhances controllability with GPT-4~\cite{openai2023gpt4}-generated guidance, and SceneDiffuser~\cite{jiang2024scenediffuser} improves inference efficiency. However, none train with direct language conditioning.

\subsection{Language in Autonomous Driving}

LLMs have enabled language integration into autonomous driving. Early datasets~\cite{yu2020bdd100k,nuscene_qa} focus on spatial annotations, while DriveVLM’s Graph VQA explores vision-language reasoning. WOMD-Reasoning~\cite{li2024womd} introduces 409K QAs on traffic-rule interactions, and ProSim-Instruct-520k~\cite{prosim} pairs 10M Llama3-70B-generated prompts with 520K scenarios from the Waymo Open Dataset. In contrast, \ourdataset is directly constructed from the interactive subset of the Waymo Open Dataset, ensuring a targeted selection of interactive scenarios. Additionally, our annotations are collected from human experts rather than LLMs, focusing specifically on high-quality interactive behavior labeling.

Language-conditioned simulation methods vary. LCTGen~\cite{lctgen} generates trajectories in an \textit{open-loop} manner based on discrete attributes, while ProSim~\cite{prosim} uses an autoregressive framework for goal-driven simulation. We adopt a diffusion-based approach for greater flexibility beyond the training distribution, enabling more adaptive and interactive simulations, as discussed in \cref{sec:safety}.

\begin{figure*}
    \centering
    \includegraphics[width=\textwidth]{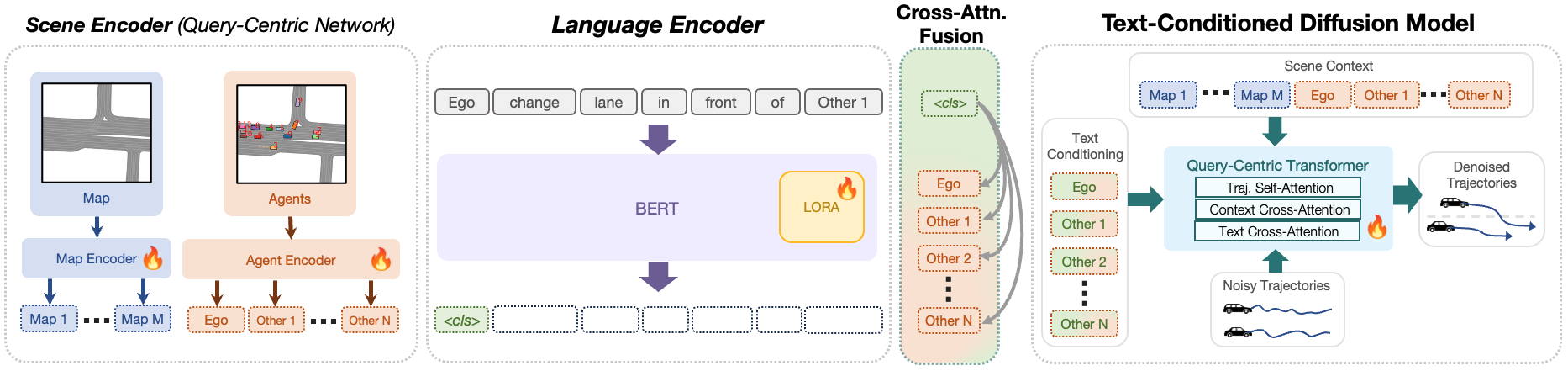} 
    \caption{\textbf{Overview of \ourmethod.} We introduce \ourmethod, a novel language-controlled diffusion-based model for trajectory simulation that incorporates HD maps, agent histories, and text descriptions, enabling behaviorally nuanced trajectory generation.}
    \label{fig:arch} 
\end{figure*}

%% file: sec/3_method.tex
\section{\ourdataset Dataset}
Natural language conditioning for traffic simulation relies heavily on the scale and quality of the underlying data. To address this, we introduce \ourdataset, a central contribution of this work, designed to capture nuanced agent-agent interactions in real-world driving contexts. Our dataset is constructed through an extensive human-labeling effort on the Waymo Motion \cite{waymo_dataset} and NuPlan \cite{caesar2021nuplan} datasets, with additional single-agent behavioral labels generated via heuristic annotations. 

Unlike previous datasets such as ProSimInstruct~\cite{prosim}, which provide general trajectory labels, we focus specifically on \textit{interactive agents}, ensuring that agent-agent interactions—such as merging, yielding, and passing—are explicitly labeled. This emphasis enables more realistic and flexible training of language-conditioned diffusion models for diverse traffic scenarios.

\subsection{Dataset Composition} 

The \ourdataset dataset includes annotations from both the Waymo \cite{ waymo_dataset} and NuPlan \cite{caesar2021nuplan} datasets, covering a diverse range of environments, agent types, and interaction scenarios. For Waymo, 125k scenes were annotated with human-generated interaction labels and 405k scenes annotated with heuristic labels. For nuPlan, 12k scenes were annotated with interaction labels and 150k scenes annotated with heuristic labels, each lasting 15 seconds. 



\subsection{Human-Labeling Process} 
To ensure high-quality and efficient interaction annotations, our labeling protocol was carefully designed to leverage human expertise in defining complex driving behaviors. Labelers were provided with top-down video representations of each scene, enabling them to accurately observe and annotate agent interactions. Using these visualizations, we developed a scalable and efficient pipeline for annotating complex interactions through a structured multi-choice question framework. The process consists of the following steps:

\begin{enumerate}
    \item \textit{Identify Interacting Agents:} For Waymo, we used the interacting agent IDs provided by the dataset. For nuPlan, we asked labelers to manually identify one agent within each scene that exhibits an interaction with the ego vehicle.
    \item \textit{Classify Interaction Type:} Labelers select the primary behavior of each interaction agent, choosing from six interaction types: (1) Lane Changing, (2) Following/Stopping Behind, (3) Yielding, (4) Passing, (5) Overtaking, and (6) Merging.
    \item \textit{Classify Interaction Subtypes:} For each chosen interaction type, labelers provide a more granular interaction subtype, capturing specifics like ``Changing lane for overtaking,'' and ``Intersection yielding''. A full list of interaction subtypes is provided in \cref*{supp:dataset} and in \cref{fig:dataset}.
\end{enumerate}
As shown in \cref{fig:dataset}, most real-world driving behaviors can be effectively captured by our carefully designed interaction types, providing a comprehensive framework for annotating complex agent interactions.

\subsection{Heuristic Annotation Process} 
In addition to interaction labels, \ourdataset contains labels for single-agent states/actions, which were generated automatically using a set of carefully calibrated heuristics. For the pedestrians and cyclists labels, we selected all that applied from the following list: static, crossing the street, walking along the road, and moving. For the vehicle labels, we selected all that apply from the following list: parked, off the main roads, static, moving slowly, speeding up, slowing down, moving at a constant speed, turning right, turning left, going straight, crossing an intersection, approaching an intersection, lane position (e.g. rightmost), changing lanes from-to (e.g. middle-to-rightmost). This enabled comprehensive behavioral modeling in scenarios with both explicit and implicit agent interactions.

Our human-annotated dataset provides rich annotations for interactive simulation and behavior studies, enabling in-depth analysis of agent interactions and unlocking future possibilities for language-to-simulation research.


\section{Problem Formulation}\label{sec:problem}

In traffic simulation, we model behaviors of \( N \) agents through a centralized function \( g_{\theta} \), enabling realistic and controllable agent actions via language-conditioned commands \( \mathbf{e}_{\text{lang}} \). The joint agent state at each timestep \( t \) is \(\mathbf{s}_t = [\mathbf{s}^1_t, \ldots, \mathbf{s}^N_t]\) with actions \(\mathbf{a}_t = [\mathbf{a}^1_t, \ldots, \mathbf{a}^N_t]\), transitioning via \( f \) based on unicycle dynamics: \(\mathbf{s}_{t+1} = f(\mathbf{s}_t, \mathbf{a}_t)\).

Each agent shares context \(\mathbf{c}_t\), including HD map \( I \) and historical states \(\mathbf{S}_{t-T_{\text{hist}}:t}\). The function \( g_{\theta} \) generates future trajectories \(\{\mathbf{s}^i_{t:t+T}\}_{i=1}^{N}\) based on \((\mathbf{c}_t, \mathbf{e}_{\text{lang}})\), trained on real-world data to ensure realism and flexibility to user-defined scenarios.

We use diffusion models to produce text-conditioned trajectories, reversing a forward noising process from real trajectories \( \tau_0 \sim q(\tau_0) \) into noisy sequences \((\tau_1, \ldots, \tau_K)\) via Gaussian noise:
\[
q(\tau_{1:K} | \tau_0) := \prod_{k=1}^{K} q(\tau_k | \tau_{k-1}),
\]
\[
q(\tau_k | \tau_{k-1}) := \mathcal{N}(\tau_k; \sqrt{1 - \beta_k}\tau_{k-1}, \beta_k\mathbf{I}).
\]
The model learns to denoise \(\tau_K\) back to \(\tau_0\), integrating text encoding \(\mathbf{e}_{\text{text}}\) to influence mean predictions:
\[
p_{\theta}(\tau_{k-1} | \tau_k, \mathbf{c}, \mathbf{e}_{\text{lang}}) := \mathcal{N}(\tau_{k-1}; \mu_\theta(\tau_k, k, \mathbf{c}, \mathbf{e}_{\text{lang}}), \Sigma_k).
\]
This enables generation of scene- and text-aligned future trajectories.


\section{\ourmethod}

We introduce \ourmethod, a scene-diffusion model designed to capture the joint distribution of interactive behaviors among all agents within multi-agent environments. To achieve this, \ourmethod comprises two key components: 1) an encoder that effectively represents the scene context, including map features and historical agent behaviors, and 2) a denoiser capable of jointly predicting future agent behaviors while providing flexibility and user-driven control via natural language inputs. \ourmethod supports trajectory generation conditioned on textual prompts and diffusion guidance, ensuring both flexibility and precise control. Additionally, we propose a novel algorithm for closed-loop training of diffusion models, further enhancing the model's performance and applicability in closed-loop simulation settings.



\subsection{Architecture}

\subsubsection{Scene Encoder}

Inspired by \cite{qcnet, shi2024mtr++}, we adopt a query-centric approach combined with Graph Neural Networks (GNNs) to model spatiotemporal relationships in the scene. A key component of this approach is the attention mechanism, which encodes relative spatial and temporal information, capturing interactions and dependencies between scene elements.  The encoder operates in a scene-independent local coordinate system. Each scene element extracts features within its own local reference frame, independent of global coordinates or the ego vehicle's position. This formulation allows for symmetric encoding across agents without being affected by variations in absolute positioning. Through this process, the encoder produces an embedding \( \mathbf{z}^i_{\text{enc}} = E_{enc}(I, \mathbf{S}_{t-T_{\text{hist}}:t}) \) for each agent \( i \).



\subsubsection{Language Encoder}

The Language Encoder module extracts agent-specific embeddings from natural language inputs and integrates them with the scene’s spatiotemporal context (see \cref{fig:arch}). This module serves two key functions: 1) providing explicit agent-specific conditioning and 2) encoding spatiotemporal data to capture agent interactions.

To achieve agent-specific conditioning, we first process the input sentence through a language model, where agent roles are explicitly \textit{rephrased} for direct conditioning. Specifically, for each agent, its role in the sentence is labeled as the “target agent,” while other agents are designated as “other agent1,” “other agent2,” and so forth. This rephrasing ensures clear distinctions in agent roles within the sentence. After processing through the language model, we obtain a sentence embedding \( \mathbf{e}_{\text{lang}} \) that captures the overall context.

We define the language encoder function \( E_{L} \) to integrate the language and scene context. Specifically, \( E_{L} \) combines the language-conditioned sentence embedding \( \mathbf{e}_{\text{lang}} \) with each agent’s context embedding \( \mathbf{z}^i_{\text{enc}} \) from the Scene Encoder:
\[
\mathbf{z}^i_{\text{lang}} = E_{L}(\mathbf{e}_{\text{lang}}, \mathbf{z}^i_{\text{enc}}),
\]
where \( \mathbf{z}^i_{\text{lang}} \) represents the final language-conditioned embedding for each agent \( i \). This embedding incorporates both the spatial relationships within the scene and the agent-specific conditioning derived from the language input.

Our design is flexible regarding the choice of language model; following the practice of \cite{shi2024yell}, we use DistillBERT \cite{sanh2019distilbert}, a smaller language encoder, to extract the $<$cls$>$ token embedding as a summary of the sentence, reducing computation cost. Empirically, we find this sufficient for language conditioning. In practice, we train the language encoder end-to-end with the diffusion model using LoRA (Low-Rank Adaptation), enabling parameter-efficient fine-tuning.


\subsubsection{Denoiser}

The denoiser consists of multiple stacked transformer blocks, each incorporating different types of attention to model complex agent behaviors and interactions. Specifically, the denoiser employs: 1) Query-centric attention across agents’ future trajectories to capture inter-agent relationships, allowing each agent to attend to other agents’ future actions; 2) Agent-Context Cross-Attention, where each agent attends to its context embedding \( \mathbf{z}^i_{\text{enc}} \) generated by the Scene Encoder, ensuring that the agent’s behavior aligns with the spatiotemporal context of the environment; and 3) Text-Cross Attention, where if a language description is provided, each agent also attends to the language-conditioned embedding \( \mathbf{z}^i_{\text{lang}} \), incorporating condition from user-defined language inputs.

\input{tbl/alg}

\subsection{Closed-loop Training}
\label{sec:cl_train}
Typically, trajectory diffusion models are trained in an open-loop fashion \cite{jiang2024scenediffuser, motiondiffuser, chang2025safesim, ctg++}. However, this mismatch between training and inference can lead to distribution shifts, where the model encounters compounding errors due to deviations from the expected trajectory distribution. A common approach to mitigate this issue during inference is to apply guidance methods or hard constraints to steer model predictions toward physically feasible behaviors. However, these techniques slow down inference speed and require careful balancing of multiple objectives, limiting adaptability in real-time simulations.

\begin{figure}[t]
    \centering
    \includegraphics[width=0.8\linewidth]{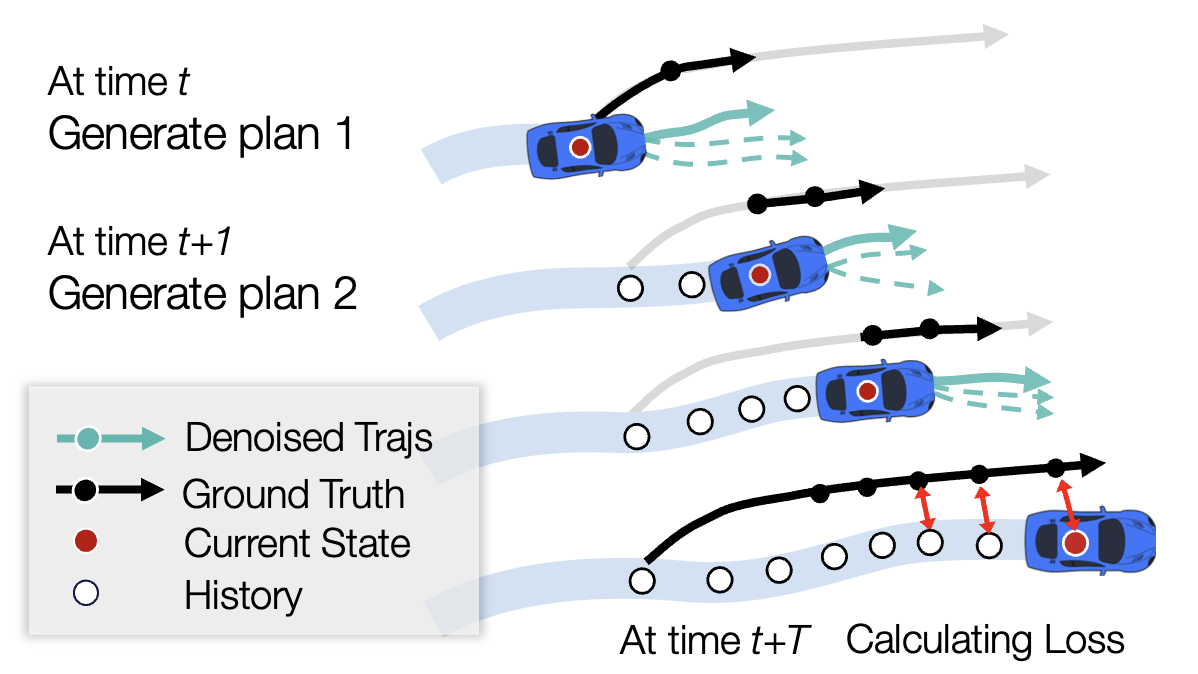} 
    \caption{\textbf{Illustration of Closed-loop Training of diffusion models.} 
    The figure demonstrates the procedure of training diffusion models in a closed-loop setting. First, the model generates multiple denoised trajectory candidates. The closest candidate to the ground truth is selected and then executed, enabling the model to experience its own distribution during training}
    \label{fig:closed_loop_training} 
\end{figure}

In contrast, closed-loop training is a widely used technique for mitigating compounding errors in closed-loop rollouts, as seen in autoregressive models \cite{prosim, suo2021trafficsim}, where past predictions are recursively used as input to better reflect real-world deployment scenarios. However, applying closed-loop training to diffusion models presents a unique challenge: the iterative denoising process involves a double for-loop, making it computationally expensive to incorporate self-generated states into training.

To address this, we propose a closed-loop training tailored for diffusion models, inspired by \cite{lin2025revisit, zhang2024closed_catk}, which better aligns the training and inference distributions without requiring multiple denoising steps during training. Instead of training with purely ground-truth trajectories, we integrate model-generated samples into training to allow the model to experience its own predictions.

As shown in \cref{fig:closed_loop_training}, our method modifies the standard diffusion training process by incorporating model-generated samples into training. At each training step, we apply forward diffusion to perturb the ground-truth trajectory segment, followed by a one-step reverse diffusion to generate multiple candidate trajectories. We then select the best candidate using a predefined distance function and execute the selected trajectory before repeating the process. The final loss is computed between the executed and ground-truth states in the global coordinate space, allowing the model to learn how to recover from accumulated errors. For a detailed breakdown of this procedure, refer to Algorithm~\ref{alg:closed_loop_diffusion}.
In practice, we employ a teacher-forcing strategy, where a subset of agents follows ground-truth states during training. This stabilizes learning and improves adherence to map-related metrics, as shown in \cref{tab:cl_ablation}.

\subsection{Controllable Behavior via Diffusion Guidance}
To enhance control over trajectory generation, we can incorporate two types of guidance: \textit{classifier-based guidance}, which optimizes differentiable objectives, and \textit{classifier-free guidance}, which blends conditional and unconditional predictions.

For classifier-based guidance, we modify the predicted mean at each denoising step using the gradient of an objective function \( J(\tau) \), steering trajectories toward desired behaviors while maintaining realism. We adopt reconstruction guidance (clean guidance) from \cite{video_diffusion, trace} to improve stability:
\begin{equation}
\tilde{\tau}_0 = \hat{\tau}_0 - \alpha \sum_k \nabla_{\tau_k} J(\hat{\tau}_0).
\end{equation}

The details of the adopted guidance functions can be found in \cref*{supp:guidance}.
For classifier-free guidance, we interpolate between conditional and unconditional predictions from \( g \):
\begin{equation}
\hat{\tau}_0 = (1 + w) \cdot g(\mathbf{e}_{\text{lang}}, \mathbf{z}_{\text{enc}}) - w \cdot g(\emptyset, \mathbf{z}_{\text{enc}})
\end{equation}
where \( g(\mathbf{e}_{\text{lang}}, \mathbf{z}_{\text{enc}}) \) includes language input \( \mathbf{e}_{\text{lang}} \), while \( g(\emptyset, \mathbf{z}_{\text{enc}}) \) omits it. The guidance weight \( w \) controls the influence of language conditioning, allowing more flexible behavior synthesis without relying on predefined objectives.

By integrating these two diffusion-based approaches, we enable more flexible and adaptable simulations, which are crucial for testing autonomous vehicles in diverse and challenging scenarios.





%% file: tbl/alg.tex
\begin{algorithm}[t]
\caption{Closed-Loop Training of Diffusion Models}
\label{alg:closed_loop_diffusion}
\begin{algorithmic}[1]

\Require Pretrained diffusion model $\theta_{\text{init}}$, dataset $\mathcal{D} = \{s_{0:T}^{(i)}\}_{i=1}^N$, distance metric $D(\cdot,\cdot)$, planning horizon $T_{\text{replan}}$, number of samples $M$, noise scale $\gamma$, denoising steps $K$
\Ensure Optimized model $\theta$

\State Initialize $\theta \gets \theta_{\text{init}}$

\For{each trajectory $s_{0:T} \sim \mathcal{D}$}    
    \For{$t = 0$ \textbf{to} $T-T_{\text{replan}}$ \textbf{step} $T_{\text{replan}}$}
        \State Extract target sequence: $s_{\text{target}} = s_{t:t+T_{\text{replan}}}$
        \State Sample noise level: $\tau \sim \mathcal{U}(1, K\gamma)$ 
        \State Add noise: $s_{\tau} = \sqrt{\alpha_\tau} s_{\text{target}} + \sqrt{1-\alpha_\tau} \epsilon$, $\epsilon \sim \mathcal{N}(0, I)$
        \State Generate $M$ denoised candidates $\hat{s}^{(1:M)}$
        \State Select $\hat{s}^{(m^*)}$ via $\arg\min_{m} D(\hat{s}^{(m)}, s_{\text{target}})$
        \State Execute $\hat{s}^{(m^*)}$, update state
    \EndFor 
    \State Compute loss $\mathcal{L} = \|\hat{s}_{0:T} - s_{0:T}\|^2$, update $\theta$
\EndFor
\State \Return $\theta$

\end{algorithmic}
\end{algorithm}

%% file: sec/4_exp.tex
\section{Experimental Results}


\input{tbl/text_cond}

\subsection{Preliminaries}
We validate our framework through experiments on real-world driving data from the Waymo Open Motion Dataset (WOMD) \cite{waymo_dataset}. We train \ourmethod on the training splits of \ourdataset and ProSim-Instruct-520k, described in \cref*{supp:prelim}, and evaluate its ability to generate behaviors that are both \textit{realistic} and \textit{controllable}.

\input{tbl/text_cond_new}

\paragraph{Metrics.}
For realism (denoted "Meta" in the tables), we follow the Waymo Open Sim Agent Challenge (WOSAC), which evaluates the distributional realism of generated trajectories \cite{waymo_sim_agent}. The challenge assesses how well joint simulation rollouts recover held-out ground truth behavior across multiple aspects, including kinematics, agent interactions, and map adherence. Further details can be found in \cref*{sup:metrics}. To evaluate \textbf{controllability}, we condition the model on ground-truth future behavior descriptions and measure the improvement in minADE compared to the unconditional model. This comparison quantifies the model’s ability to align generated behaviors with user-specified input when provided with behavior descriptions.

\input{tbl/wosac}

\subsection{Evaluation of \textbf{\ourdataset}}
We train separate instantiations of \ourmethod on the training sets of \ourdataset (InterDrive) and ProSim-Instruct-520k, performing closed-loop evaluations with and without text conditioning. As shown in \cref{tab:text_cond_new}, language conditioning on \ourdataset yields a 13.6\% reduction in minADE while maintaining a meta-realism score of 0.72, indicating that our language annotations provide useful supervisory signals. A similar improvement is observed on ProSim-Instruct-520k (12.8\% minADE reduction). Notably, \ourdataset focuses on interactive agent-agent behaviors, whereas ProSim-Instruct primarily consists of single-agent maneuvers.


\input{tbl/reb_ctg++}

\subsection{LLM-based Guidance v.s Direct Conditioning}
In \cref{tab:reb_ctg}, we compare our proposed direct language conditioning method with the LLM-based guidance approach from CTG++~\cite{ctg++}, which uses GPT to generate differentiable code-based guidance functions from language prompts (see \cref*{supp:ctg++} for details). On \ourdataset scenarios, we find that LLM-based guidance leads to worse alignment with the text prompt—even underperforming the no-text baseline. In contrast, direct conditioning produces behaviors more aligned with the ground truth (13.6\% reduction) while preserving simulation realism. Direct conditioning also outperforms LLM-based guidance in counterfactual settings, as discussed in \cref{sec:safety}. While LLM-based guidance is effective for enforcing constraints like collision avoidance and on-road driving \cite{ctg++}, it still struggles with interaction-level behavior descriptions, underscoring the benefit of direct language grounding.


\subsection{Evaluation of \ourmethod}

To enable a fair comparison with ProSim, we train \ourmethod on the ProSim-Instruct-520k training set and evaluate its performance against the publicly released ProSim model, which is trained on the same dataset. As shown in \cref{tab:text_cond_new}, both methods achieve comparable performance in simulation realism and gain in terms of minADE.  However, \ourmethod employs diffusion-based modeling, which unlocks inference-time guidance and controlled sampling, offering greater flexibility beyond the training distribution that autoregressive-based methods like Prosim does not have. We discuss these advantages in \cref{sec:safety}.

Beyond instruction-following tasks, we evaluate \ourmethod on the Waymo Sim Agents Challenge Benchmark (unconditioned simulation), with results presented in \cref{tab:wosac}. Compared to state-of-the-art diffusion models, \ourmethod performs competitively among the best diffusion-based approaches, such as VBD \cite{huang2024versatile} and SceneDiffuser \cite{jiang2024scenediffuser}.



\subsection{Text-Conditioned Safety-Critical Simulation}
\label{sec:safety}


We demonstrate \ourmethod's ability to extend beyond the training distribution by generating safety-critical scenarios conditioned on text inputs using guided sampling, as illustrated in \cref{tab:collision} and \cref{teaser}. Given the focus on collision-prone situations, standard interactive metrics are less informative; thus, we instead evaluate simulation quality using kinematics and map adherence metrics. Our findings indicate that \ourmethod successfully generates realistic traffic scenarios as well as targeted, rare, safety-critical events, achieving collision rates as high as 40\%. Compared to LLM-based guidance, \ourmethod achieves a 10\% higher collision rate and outperforms CTG++ across all metrics in safety-critical situations.
Additionally, integrating direct text conditioning enhances map adherence even under collision-focused guidance, highlighting a promising approach that combines explicit guidance with textual inputs to improve scenario diversity and simulation control. Qualitative examples are provided in the supplementary videos.

\input{tbl/collision}

\subsection{Closed-loop Training}\label{sec:exp:cl_train}

We analyze the effects of denoising steps, closed-loop training, and teacher forcing in \cref{tab:cl_ablation}. First, we find that reducing denoising steps from $K=100$ to $K=5$ maintains realism metrics, suggesting that trajectory simulation can be potentially effectively learned with fewer denoising steps, reducing computational cost. Next, introducing closed-loop training with teacher forcing improves both map adherence (0.72 → 0.79) and meta realism (0.68 → 0.70). 
Note that we also observe vehicles tend to slow down and drive off of the road when trained in a closed-loop manner without teacher forcing, despite achieving similar realism scores. This suggests that a portion of agents follow the ground-truth trajectory helps stabilize training, preventing the model from drifting into unrealistic behaviors

\input{tbl/closed_loop_ablation}

%% file: tbl/text_cond.tex


\begin{table}[t]
\centering
\resizebox{1.0\linewidth}{!}{%

\begin{tabular}{lc|cccc|c} 
\toprule
\textbf{Datset} & \textbf{Text} & \textbf{Meta $\uparrow$} & \textbf{Kinematic $\uparrow$} & \textbf{Interactive $\uparrow$} & \textbf{Map $\uparrow$} & \textbf{mADE $\downarrow$} \\ 
\midrule
\multirow{2}{*}{\ourdataset} & $\times$ & 0.72 & 0.41  & 0.80 & 0.80  & 2.65 \\
& \checkmark & 0.72 & 0.42  & 0.80 & 0.79 & 2.29 \\
\midrule
\multirow{2}{*}{ProSim-Instruct} & $\times$ & 0.72 & 0.42  & 0.79 & 0.79 & 2.89 \\
& \checkmark & 0.72 & 0.43 & 0.80 & 0.78 & 2.52 \\
\bottomrule
\end{tabular}
}
\caption{\textbf{Evaluation of \ourdataset:} We train \ourmethod separately on the \ourdataset and ProSim-Instruct-520k training sets and evaluate its performance on their respective test sets.}
\label{tab:text_cond_new}
\end{table}



%% file: tbl/text_cond_new.tex
\begin{table}[t]
\centering
\resizebox{1.0\linewidth}{!}{%
\begin{tabular}{lc|cccc|c} 
\toprule
\textbf{Method} & \textbf{Text} & \textbf{Meta $\uparrow$} & \textbf{Kinematic $\uparrow$} & \textbf{Interactive $\uparrow$} & \textbf{Map $\uparrow$} & \textbf{mADE $\downarrow$} \\ 
\midrule
\multirow{2}{*}{\ourmethod} & $\times$ & 0.72 & 0.42  & 0.79 & 0.79 & 2.89 \\
& \checkmark & 0.72 & 0.43 & 0.80 & 0.78 & 2.52 \\
\midrule
\multirow{2}{*}{ProSim} & $\times$ & 0.69 & 0.42  & 0.73 & 0.80 & 2.73 \\
& \checkmark & 0.69 & 0.42  & 0.73 & 0.81 & 2.35 \\
\bottomrule
\end{tabular}
}
\caption{\textbf{Evaluation of \ourmethod.} Closed-loop evaluation on the ProSim-Instruct-520k test set, with and without text conditioning. We compare the performance of our method (\ourmethod) against the publicly released ProSim model, both trained on the ProSim-Instruct-520k training set, ensuring a fair evaluation.}
\label{tab:text_cond_new}
\end{table}

%% file: tbl/wosac.tex
\begin{table}[t]
\centering
\resizebox{\linewidth}{!}{%

\begin{tabular}{l|cccc} 
\toprule
\textbf{Method} & \textbf{Meta $\uparrow$} & \textbf{Kinematic $\uparrow$} & \textbf{Interactive $\uparrow$} & \textbf{Map $\uparrow$} \\
\midrule
UniMM \cite{lin2025revisit} & 0.769 & 0.491 & 0.811 & 0.874  \\
SMART-tiny-CLSFT \cite{zhang2024closed} & 0.762 & 0.458 & 0.811 & 0.872 \\
\textcolor{blue}{VBD} \cite{huang2024versatile}  & 0.720 & 0.417 & 0.814 & 0.776 \\
\textcolor{blue}{\ourmethod} & 0.719 & 0.426 & 0.795 & 0.789 \\
ProSim \cite{prosim} & 0.718 & 0.401 & 0.778 & 0.822 \\
\textcolor{blue}{SceneDiffuser} \cite{jiang2024scenediffuser} & 0.703 & 0.430 & 0.776 & 0.768 \\
\textcolor{blue}{SceneDMF} \cite{guo2023scenedm}  & 0.628 & 0.371 & 0.683 & 0.703 \\
\bottomrule
\multicolumn{5}{r}{\footnotesize \textbf{\textcolor{blue}{Blue:} Diffusion-Based Methods}\vspace{-.5em}} 
\end{tabular}
}
\caption{\textbf{Results on WOSAC Test Set.} \ourmethod performs competitively among the best diffusion-based approaches, such as VBD \cite{huang2024versatile} and SceneDiffuser \cite{jiang2024scenediffuser}.}
\label{tab:wosac}
\end{table}

%% file: tbl/reb_ctg++.tex
\begin{table}[t]
\centering
\resizebox{1.0\linewidth}{!}{%
\begin{tabular}{l|cccc|c} 
\toprule
\textbf{Text Conditioning} & \textbf{Meta $\uparrow$} & \textbf{Kinematic $\uparrow$} & \textbf{Interactive $\uparrow$} & \textbf{Map $\uparrow$} & \textbf{mADE $\downarrow$} \\ 
\midrule
None          & 0.72 & 0.41 & 0.80 & 0.80 & 2.65 \\
Direct Condition (Ours)& 0.72 & 0.42 & 0.80 & 0.79 & 2.29 \\
{LLM-Based Guidance (CTG++ Style)} & {0.70} & {0.42} & {0.75} & {0.80} & {2.70} \\\bottomrule
\end{tabular}
}
\vspace{-.8em}
\caption{\textbf{Text Conditioning Evaluation.} We compare our proposed direct text conditioning method to the LLM-based guidance method proposed by CTG++ ~\cite{ctg++}. \vspace{-.8em}}
\label{tab:reb_ctg}
\end{table}

%% file: tbl/collision.tex
\begin{table}[t]
\centering
\resizebox{\linewidth}{!}{%
\begin{tabular}{l|cc|ccc} 
\multicolumn{1}{l}{} & \multicolumn{2}{c}{\textbf{COLLISION}} & \multicolumn{1}{c}{} & \multicolumn{1}{c}{} &  \multicolumn{1}{c}{} \\
\toprule
\textbf{Text Conditioning} & 
\textbf{Guidance} & \textbf{Rate} $\uparrow$ & 
\textbf{Kinematic} $\uparrow$ & 
\textbf{Map} $\uparrow$ & 
\textbf{mADE} $\downarrow$ \\
\midrule
None & $\times$ & 0.04 & 0.42 & 0.81 & 3.04 \\
\midrule
None & $\checkmark$ & 0.41 & 0.39 & 0.70 & 4.93 \\
Direct Condition (Ours) & $\checkmark$ & \textbf{0.43} & \textbf{0.41} & \textbf{0.74} & \textbf{4.43} \\
{LLM-based Guidance (CTG++ Style)} & $\checkmark$ & {0.33} & {0.37} & {0.72} & {4.67} \\
\bottomrule
\end{tabular}
}
\caption{\textbf{Text-Conditioned Safety-Critical Simulation with \ourmethod.} We demonstrate \ourmethod's ability to extend beyond the training distribution by generating safety-critical scenarios conditioned on text inputs using guided sampling.}
\label{tab:collision}
\end{table}

%% file: tbl/closed_loop_ablation.tex

\begin{table}[t]
\centering
\resizebox{\linewidth}{!}{%
\begin{tabular}{l|cccc} 
\toprule
\textbf{Setting} & \textbf{Meta $\uparrow$} & \textbf{Kinematic $\uparrow$} & \textbf{Interactive $\uparrow$} & \textbf{Map $\uparrow$} \\ 
\midrule
Open-loop (K=100) & 0.68 & 0.42 & 0.77 & 0.73 \\
Open-loop (K=5) & 0.68 & 0.41 & 0.78 & 0.72 \\
Closed-loop & 0.69 & 0.38 & 0.78 & 0.75 \\
Closed-loop w/ Teacher & 0.70 & 0.39 & 0.78 & 0.79 \\
\bottomrule
\end{tabular}
}
\caption{\textbf{Ablation Study on Closed-Loop Training.} Study conducted on validation set of WOSAC challenge.}
\label{tab:cl_ablation}
\end{table}

%% file: sec/5_conc.tex
\section{Conclusion}  

\ourmethod advances autonomous vehicle simulation by leveraging language-conditioned diffusion models to generate diverse, behaviorally rich scenarios. Unlike prior works, it supports direct language conditioning for intuitive behavior specification and guidance-based control for counterfactual and targeted scenario generation. This flexibility makes \ourmethod well-suited for scalable, controllable AV simulation, enabling safety-critical testing and interactive scenario design.  

A key component of our approach is \ourdataset, which focuses on interactive agents, providing rich human-labeled and heuristic annotations that enhance realism and diversity. By prioritizing agent-agent interactions, \ourdataset strengthens training for language-conditioned models. Empirical results on the Waymo Motion Dataset show that \ourmethod generates realistic, language-aligned trajectories while preserving simulation realism. In summary, \ourmethod demonstrates the potential of language-conditioned diffusion models for AV simulation by combining natural language with guidance-based control. 





%% file: sec/ack.tex
\clearpage

\section*{Acknowledgements}

This work was part of W.J. Chang’s summer internship at NEC Labs America, and he is also supported by the National Science Foundation Graduate Research Fellowship Program under Grant No. DGE 2146752. Any opinions, findings, and conclusions or recommendations expressed in this material are those of the author(s) and do not necessarily reflect the views of the National Science Foundation. This study was funded in part by the InnoHK initiative of the Innovation and Technology Commission of the Hong Kong Special Administrative Region Government via the Hong Kong Centre for Logistics Robotics. The authors would like to thank Yichen Xie and Rian Tian for their insightful discussions, and Chih-Ling Chang for her helpful suggestions and assistance with figures and presentations.

%% file: sec/X_suppl.tex
\clearpage
\setcounter{page}{1}
\maketitlesupplementary
\appendix
\renewcommand{\thesection}{\Alph{section}}

\section{Experimental Details}\label{supp:prelim}

\subsection{\textbf{\ourdataset}}

The training split of \ourdataset includes 100k human-annotated language-trajectory pairs and 405k pairs annotated heuristically. The test split of \ourdataset contains 25k prompt-scenario pairs with both human and heuristic annotations. To address open-set language input, we augment \ourdataset's categorical annotations using GPT-4 \cite{openai2023gpt4} to generate approximately 20 rephrasings for each annotated behavior. This augmentation expands the range of language variations the model encounters, improving its robustness to diverse user inputs.

During training, we also use \textit{compositional} prompts that combine agent-agent interaction descriptions with heuristic action labels (e.g., \textit{speed up}, \textit{turn left}, \textit{wait}) into unified instructions. This compositionality supports the flexibility of behavior expression, and we apply it consistently across both training and evaluation. Note that our instructions do not include explicit temporal conditioning, which we leave for future work.

\subsection{ProSim-Instruct-520k}

ProSim-Instruct-520k is a multimodal dataset designed for promptable traffic simulation, containing over 10 million text prompts paired with 520,000 driving scenarios. Each scenario includes goal points, route sketches, action tags describing agent behaviors, and text instructions generated by Llama3-70B. In contrast, \ourdataset is directly constructed from the interactive subset of the Waymo Open Dataset, ensuring a targeted selection of interactive scenarios. Additionally, our annotations are collected from human experts rather than LLMs, focusing specifically on high-quality interactive behavior labeling.

\subsection{LLM-Based Guidance Details}\label{supp:ctg++}

In this section, we provide implementation details for LLM-based guidance conditioning (CTG++ style~\cite{ctg++}) method in \cref{tab:reb_ctg}, which generates differentiable loss functions conditioned on text descriptions. We use the o3-mini model via OpenAI APIs to generate loss functions that guide vehicle behaviors, following the \href{https://github.com/NVlabs/CTG}{implementation}. The method leverages the same backbone and weights for all experiments to ensure consistency.

Since \ourdataset uses a fixed vocabulary, we generate a unique function for each interactive description and heuristic, and combine them by scaling the loss to a common range. LLM-based guidance may not work in the first iteration, as the generated functions often contain errors or inconsistencies. Common failure cases include issues with array shape mismatches, map-related functions, and the assumption of unseen functions. To address this, we manually correct the generated functions by providing more specific instructions to GPT. This process typically requires 3-5 cycles to refine the guidance functions and ensure there are no compilation errors.


\subsection{Testing Subsets}
We evaluate all experiments on a 2\% subset of the data, consisting of approximately 1,100 scenarios. Specifically, \ourdataset is tested on the validation interactive subset of the Waymo dataset, while ProSim-Instruct-520k is evaluated on the validation subset. 

\subsection{WOSAC Challenge Metrics}\label{sup:metrics}

The Waymo Open Sim Agent Challenge (WOSAC) evaluates simulation quality by computing negative log-likelihood (NLL) scores over nine predefined statistical features, covering kinematics, agent interactions, and map adherence, where each feature is evaluated independently. The challenge requires simulating up to 128 agents per scene for 8 seconds, generating 32 joint agents future samples per scenario. The negative log-likelihood is then computed based on an approximate empirical distribution constructed from the simulated trajectories.

For a given scenario \( i \) and target agent \( a \), the likelihood of the true trajectory under the empirical distribution of simulated samples is given by:
\begin{equation}
\text{NLL}(i, a, t, j) = -\log p_{i,j,a}(F_j(x^*(i, a, t)))
\end{equation}
where \( p_{i,j,a}(.) \) is the empirical histogram distribution of statistic \( F_j \) obtained from the generated samples. A lower NLL indicates that the simulated distribution closely matches real-world behavior.

To obtain a per-scenario metric, the NLL values are aggregated over all valid timesteps:
\begin{equation}
\resizebox{0.9\linewidth}{!}{$
m(a, i, j) = \exp\left(-\left[ \frac{1}{N(i, a)} \sum_{t} v(i, a, t) \text{NLL}(i, a, t, j) \right] \right)
$}
\end{equation}
where \( N(i, a) = \sum_{t} v(i, a, t) \) represents the number of valid timesteps for target agent \( a \). The final scenario-level metric is then computed by averaging over the target agents:
\begin{equation}
m(i, j) = \frac{1}{A_{\text{target}}} \sum_{a} m(a, i, j)
\end{equation}
where \( A_{\text{target}} \) represents the number of target agents.

To adapt WOSAC for language-conditioned interactive driving, we focus on evaluating target agents that have explicit interactive descriptions in natural language on \ourdataset, which is drawn from the validation interactive set, referred to as the objects of interest field in the Waymo Open Dataset format \cite{waymo_dataset}. 

For ProsimInstruct evaluation, since the annotations are constructed from the validation set and may not contain objects of interest labels, we follow the original target agents from the validation set.

To compute the final composite metric for ranking submissions, WOSAC takes a weighted average over all component metrics:
\begin{equation}
M_K = \frac{1}{N} \sum_{i=1}^{N} \sum_{j=1}^{M} w_j m_K(i, j), \quad \sum_{j=1}^{M} w_j = 1
\end{equation}
where \( M = 9 \) represents the nine statistical features, and \( w_j \) are manually assigned weights for each metric. We detail the definitions of the component metrics below:

\noindent\textbf{Kinematic Metrics:} 
\begin{itemize}
    \item \textbf{Linear Speed:} Measures the magnitude of the first derivative of position, \( \| v \| = \left\| \frac{x_{t+1} - x_t}{\Delta t} \right\|_2 \), reflecting the agent's velocity in 3D space.
    \item \textbf{Linear Acceleration Magnitude:} Represents the magnitude of the second derivative of position, \( \frac{\| v_{t+1} - v_t \|}{\Delta t} \), describing the agent's acceleration.
    \item \textbf{Angular Speed:} Calculates the rate of change of the agent's heading, \( \omega = \frac{d(\theta_{t+1}, \theta_t)}{\Delta t} \), where \( d(\cdot) \) is the minimal angular difference on the unit circle.
    \item \textbf{Angular Acceleration Magnitude:} Measures the rate of change of angular speed, \( \frac{d(\omega_{t+1}, \omega_t)}{\Delta t} \).
\end{itemize}

\noindent\textbf{Interaction Metrics:}
\begin{itemize}
    \item \textbf{Distance to Nearest Object:} The signed distance to the nearest object in the scene, calculated using the GJK distance algorithm.
    \item \textbf{Collisions:} Detected when the signed distance to the nearest object becomes negative, indicating that two objects have collided.
    \item \textbf{Time-to-Collision (TTC):} Estimates the time before a collision occurs, assuming constant velocities.
\end{itemize}

\noindent\textbf{Map Metric:}
\begin{itemize}
    \item \textbf{Distance to Road Edge:} The signed distance to the nearest road edge in the scene.
    \item \textbf{Road Departure:} Indicates whether an agent has left the road at any point in time, based on the signed distance to the road edge.
\end{itemize}

For more details on each metric, refer to the original WOSAC Challenge paper \cite{waymo_sim_agent}.

\input{tbl/cfg}



\section{Implementation Details}\label{supp:impl}

\subsection{Training Details}

To maximize use of the human annotations in \ourdataset, we apply a biased sampling strategy to balance the training data. Specifically, we upsample human-annotated samples to represent 50\% of each training batch and include 30\% of heuristic descriptions in a given scene per sample. This approach allows for the simultaneous training of language-conditioned and unconditional diffusion models, optimizing both modes within the framework.

The training process for \ourmethod consists of two stages. In the first stage, the scene encoder and diffusion model are trained without text conditioning for 60,000 iterations using a batch size of 32. In the second stage, the scene encoder, language encoder, and diffusion model are trained with text conditioning for an additional 20,000 iterations using a batch size of 2048. The language encoder is implemented with a LoRA module \cite{hu2021lora}, which updates only the linear projection layers of the query and key matrices in DistillBERT \cite{sanh2019distilbert}, configured with \( R = 16 \) and \( \alpha = 0.4 \). Both stages employ the Adam optimizer \cite{kingma2014adam} with an initial learning rate of \( 1 \times 10^{-3} \). The diffusion model implementation follows methodologies from open-source repositories \cite{ivanovic2023trajdata, CTG}.

To distill our pretrained scene-diffusion model from \(K=100\) to \(K=5\), we train the model using a new denoising schedule for 20,000 iterations with a batch size of 256. The original pretrained scene-diffusion model has a prediction horizon of \(T=16\) with a frequency of 0.5 Hz.

\subsection{Closed-loop Training Details}

For closed-loop training of diffusion models, we first pre-train our scene-diffusion model with \(K=100\) denoising steps. We then distill the denoiser to \(K=5\) steps before applying closed-loop training, as described in \cref*{sec:cl_train}.

We modify the model's prediction horizon from \(T=16\) to \(T=8\) under 2 Hz to accommodate multi-step unrolling with ground truth actions, using a replanning interval of \(T_{\text{replan}}=2\). Given 16 steps of ground truth future trajectories, we can perform four iterations of closed-loop training. During this process, the best sample among \(M=8\) candidates is selected for execution, and the adopted forward diffusion ratio $\gamma=0.6$.

The teacher-forcing ratio is set to 50\%. When applied, 70\% of agents are randomly sampled to follow the ground truth states throughout the unrolling process. The model is trained with a learning rate of \(1 \times 10^{-5}\) using an effective batch size of 32 for around 50,000 iterations, which takes around 12 hours on 8 8×A6000 GPUs and 32 CPU cores.

Additionally, we incorporate an auxiliary non-collision loss from \cref{supp:collision} with a relative weighting of 0.1. To enhance robustness, we randomly drop text conditioning and agent history with a probability of 50\% during training.
\subsection{Inference Frequency}
Per-sample inference with 5 de-noising steps takes 4.66±0.06 seconds per 8-second simulation (1 Hz replan) using 1×A6000 GPU and 4 CPU cores.Speedups via map caching, parallelism, and distillation can 069
be adopted for scalability.

\subsection{Denoising Process}
At each denoising step, the model predicts the mean of the next denoised action trajectory. Instead of predicting the noise \( \epsilon \) used to corrupt the trajectory, the model directly outputs the clean denoised trajectory \( \hat{\tau}_0 \). The predicted mean for reconstructing \( \tau_{k-1} \) from \( \tau_k \) is defined as:

\begin{equation}
\resizebox{0.9\linewidth}{!}{$
\tau_{k-1} = \mu_\theta(\tau_k, \hat{\tau}_0) = 
\frac{\sqrt{\bar{\alpha}_{k-1}}\beta_k}{1 - \bar{\alpha}_k}\hat{\tau}_0 + 
\frac{\sqrt{\alpha_k}(1 - \bar{\alpha}_{k-1})}{1 - \bar{\alpha}_k}\tau_k,
$}
\end{equation}

Where \( \beta_k \) represents the variance from the noise schedule in the diffusion process, \( \alpha_k = 1 - \beta_k \) denotes the incremental noise reduction at each step, and \( \bar{\alpha}_k = \prod_{j=0}^{k} \alpha_j \) is the cumulative product of \( \alpha_j \) up to step \( k \).

\subsection{Diffusion Process and Inference Details}
For the diffusion process, we utilize a cosine variance schedule, with the number of diffusion steps set to \( K = 100 \) for pretrained diffusion model and $K=5$ for the closed-loop trained diffusion models. The cosine variance scheduler followed \cite{cosine_schedule}, with $s=0.008$. The model operates on a 1.1-second trajectory history and is trained to predict the next 8.0 seconds with a time step \( \Delta t = 0.5 \). During inference, we use a DDIM sampler \cite{ddim} with a stride step 1 during inference. During inference, we sample $M=64$ joint future samples for all agent, and only select the one joint agent sample with lowest collision loss. The \textbf{per-sample inference} time is 4.66±0.06 seconds for an 8-second simulation, with a 1 Hz replan rate. This process utilizes a 1×A6000 GPU and 4 CPU cores. To improve scalability, speedups can be achieved through techniques such as map caching, parallelism, and distillation.

\section{Discussion and Limitation}
While the model generally follows instructions well, we observe that failure cases often involve conflicts between language input and recent history (e.g., past 1s of motion). In such cases, the model tends to prioritize history, leading to behavior misaligned with the instruction. This is particularly noticeable when the vehicle is static, making conditioning signals harder to take effect. Additionally, minADE may not fully capture instruction adherence; future work could explore human evaluations to better assess language-action alignment.


\section{Guidance Details}\label{supp:guidance}

\subsection{Classifier-Free Guidance}\label{sec:exp:cfg}

Contrary to previous diffusion models that rely on classifier guidance, direct conditioning enables control through the learned data distribution rather than predefined objectives. Classifier-free guidance leverages this distribution for generation without requiring domain-specific priors.

As shown in \cref{tab:cfg}, we observe that moderate classifier-free guidance (weights 0.0–1.0) maintains realism and controllability, while higher weights (1.0) degrade map adherence (0.81 → 0.73) and interactive realism (0.79 → 0.76), suggesting that excessive text conditioning misaligns trajectories with the map structure. Qualitatively, while stronger guidance improves instruction-following, it also tends to produce more off-road samples.

\subsection{Collision Guidance}
We define the collision cost as
$$
J_{\text{coll}} = -\sum_{t=1}^{T} d(t),
$$
where d(t) is the distance between the adversarial and target agents at time step t over the planning horizon T. Minimizing $J_{\text{coll}}$ encourages collisions. In \cref{tab:collision}, one of the interactive agent from the \ourdataset is designated as adversarial and the other as the target. Note that, we only compute the gradient with respect to the adversarial agent.


\subsection{Non-Collision Guidance}
\label{supp:collision}

To detect collisions in a differentiable way, we approximate each agent \(i\) with \(D\) equally spaced disks of radius \(r_i\) \cite{suo2021trafficsim}. For any pair of agents \((i, j)\), let \(d\) be the minimal distance between their respective disk centers at time \(t\). If \(d\) is less than the sum of their radii, the circles overlap. Formally, the pairwise collision loss for agents \(i\) and \(j\) at time \(\tau\) is:

\begin{equation}
\resizebox{0.9\linewidth}{!}{$
J_{\text{pair}}\bigl(\tau_i,\,\tau_j\bigr)
=
\begin{cases}
1 - \dfrac{d}{r_i + r_j}, & \text{if } d \le r_i + r_j,\\[6pt]
0, & \text{otherwise}.
\end{cases}
$}
\end{equation}

We sum over all agent pairs and all timesteps \(t = 0, \dots, T\) to obtain the total collision loss:

\begin{equation}
\resizebox{0.9\linewidth}{!}{$
J_{\text{no collision}}
=
\frac{1}{N^2}
\sum_{i \neq j}
\max\!\Bigl(1,\,
\sum_{\tau = 0}^{T}
J_{\text{pair}}\bigl(\tau_i,\,\tau_j\bigr)\Bigr),    
$}
\end{equation}

where \(N\) is the total number of agents, \(T\) is the planning horizon, and \(\tau_i\) represents the state of agent \(i\) at time \(\tau\). If no disks overlap, \(J_{\text{pair}}\) is zero; fully overlapping disks produce a maximum penalty of \(1\).

\section{\ourdataset Interaction Type Definitions} \label{supp:dataset}

This section defines the detailed behaviors considered in our study and the corresponding labeling process used to categorize them.

\ourdataset uses a scalable human labeling process based on multiple-choice questions organized into large categories and subcategories to define and categorize behaviors efficiently. This structured approach reduces ambiguity, provides clear guidance for annotators, and ensures consistent, high-quality labels. By leveraging this method, \ourdataset achieves 2.5 times the size of the WOMD-Reasoning dataset, with higher-quality annotations and slightly lower overall labeling costs.

In addition to categorizing interaction types, annotators also identify whether a pair of agents is interacting. This process accounts for the possibility of \textit{asymmetric interactions}, where one agent interacts with another, but the reverse may not be true. For example, Agent 2 may adjust its behavior in response to Agent 1 (e.g., yielding or avoiding), while Agent 1 may proceed unaffected, exhibiting no interaction.

\begin{itemize}
    \item \textbf{Lane Change:} Lane change interactions involve moving from one lane to another for various purposes:
    \begin{itemize}
        \item \textbf{Changing lane for turn or exit:} Moving to another lane in preparation for turning or exiting.
        \item \textbf{Changing lane for overtaking:} Moving to another lane to pass a slower vehicle.
        \item \textbf{Lane-change for avoiding obstacles or slower traffic:} Changing lanes to bypass road obstacles or slower-moving vehicles.
        \item \textbf{Lane-change for merging:} Changing lanes to merge into another stream of traffic.
        \item \textbf{Changing lane with lead or trail:} Performing a lane change with another vehicle directly ahead (lead) or behind (trail), requiring extra caution.
    \end{itemize}

    \item \textbf{Following/Stopping Behind:} These interactions involve adjusting speed and distance while following or stopping behind another vehicle:
    \begin{itemize}
        \item \textbf{Following with a lead vehicle:} Adjusting speed to maintain a safe distance while following another vehicle.
        \item \textbf{Following a slow-moving lead:} Driving slower than desired to follow a slower vehicle ahead.
        \item \textbf{Tailgating:} Driving too closely behind another vehicle, often considered aggressive driving.
        \item \textbf{Stopping behind a lead vehicle:} Coming to a stop behind another vehicle, typically at traffic lights or stop signs.
        \item \textbf{Stopping behind an intersection:} Coming to a stop before entering an intersection.
    \end{itemize}

    \item \textbf{Yielding:} Yielding involves giving the right of way to other road users in specific situations:
    \begin{itemize}
        \item \textbf{Intersection yielding:} Yielding to oncoming traffic or other road users at intersections.
        \item \textbf{Yielding before merging or lane-change:} Yielding to ongoing traffic when changing a lane or merging.
        \item \textbf{Yielding to merging or lane-change cars:} Yielding to cars that change lanes or merge into the current lane.
        \item \textbf{Waiting for a pedestrian to cross:} Yielding to a pedestrian at a crosswalk or intersection, allowing them to cross safely.
        \item \textbf{Roundabout yielding:} Yielding to vehicles already in a roundabout before entering.
        \item \textbf{Pedestrian yielding to vehicles:} Pedestrians pause and give way to oncoming vehicles before crossing the road.
    \end{itemize}

    \item \textbf{Passing:} Passing involves moving past vehicles, pedestrians, or obstacles without yielding:
    \begin{itemize}
        \item \textbf{Passing through an intersection with yielding vehicles:} Moving through an intersection without yielding, while other cars yield.
        \item \textbf{Passing a pedestrian:} Moving past a pedestrian walking near the road or on a crosswalk, ensuring a safe distance.
        \item \textbf{Pedestrian passing a vehicle:} A pedestrian moves around or crosses in front of a stationary vehicle.
        \item \textbf{Passing through a roundabout:} Navigating through a roundabout without stopping, maintaining the right of way.
        \item \textbf{Maintaining speed while driving:} Driving straight or turning while maintaining the original speed without yielding or merging.
        \item \textbf{Passing as a leading vehicle:} Passing with other vehicles following.
        \item \textbf{Pedestrian or cyclist crossing the road:} Pedestrians or cyclists crossing the road, usually with vehicles yielding.
    \end{itemize}

    \item \textbf{Overtaking:} Overtaking involves actively moving ahead of another vehicle:
    \begin{itemize}
        \item \textbf{Car avoidance:} Taking evasive action to avoid another vehicle, often involving swerving, braking, or accelerating.
        \item \textbf{Standard overtaking:} Passing a slower vehicle by moving to an adjacent lane and returning to the original lane.
        \item \textbf{High-speed overtaking:} Passing at higher speeds on highways, requiring careful attention to speed and distance.
    \end{itemize}

    \item \textbf{Merging:} Merging involves entering a lane of traffic from a merging lane, on-ramp, or after a lane reduction:
    \begin{itemize}
        \item \textbf{Standard merge:} Entering the flow of traffic from a merging lane or on-ramp.
        \item \textbf{Lane reduction merge:} Merging into an adjacent lane when a lane ends due to road conditions.
        \item \textbf{Zipper merge:} A coordinated merge where vehicles in two lanes alternate into a single lane.
        \item \textbf{Highway on-ramp accelerating merge:} Entering a highway while accelerating to match traffic speed.
        \item \textbf{Late merge:} Merging closer to the end of the merging lane in congested traffic.
    \end{itemize}

    \item \textbf{Other:}
    \begin{itemize}
        \item \textbf{Undefined behavior:} For scenarios where a specific behavior type does not exist in the predefined options. In this case, the human labeler will type in behavior descriptions. 
        \item \textbf{No interaction:} For cases where no interaction occurs.
        \item \textbf{Unknown status:} For scenarios where the behavior cannot be determined due to insufficient or unclear data.
    \end{itemize}
\end{itemize}



%% file: tbl/cfg.tex

\begin{table}[t]
\centering
\resizebox{\linewidth}{!}{%
\begin{tabular}{l|cccc|c} 
\toprule
 \textbf{CFG Weight} & \textbf{Meta $\uparrow$} & \textbf{Kinematic $\uparrow$} & \textbf{Interactive $\uparrow$} & \textbf{Map $\uparrow$} & \textbf{mADE $\downarrow$} \\ 
\midrule
-1.0  & 0.72 & 0.40 & 0.79 & 0.81 & 3.38 \\
0.0   & 0.72 & 0.41 & 0.79 & 0.81 & 2.90 \\
0.5   & 0.71 & 0.41 & 0.79 & 0.78 & 2.90 \\
1.0   & 0.70 & 0.41 & 0.78 & 0.77 & 2.97 \\
2.0   & 0.68 & 0.40 & 0.76 & 0.73 & 3.21 \\
\bottomrule
\end{tabular}
}
\caption{\textbf{Analysis of Text Conditioning Strength.} We evaluate the impact of text conditioning in \ourmethod by varying the classifier-free guidance (CFG) weight. CFG=-1.0 represents the unconditional setting.}
\label{tab:cfg}
\end{table}